# Advances in Natural Language Question Answering: A Review

K.S.D. Ishwari[#1], A.K.R.R. Aneeze[#2], S. Sudheesan[#3], H.J.D.A. Karunaratne[#*4], A. Nugaliyadde[#5], Y. Mallawarrachchi[#6]

[#]*Faculty of Computing, Sri Lanka Institute of Information Technology*
*Sri Lanka*

[*]*College of Science, Health, Engineering and Education,*
*Australia*

[1]it15067098@my.sliit.lk
[2]it15060372@my.sliit.lk
[3]it15109668@my.sliit.lk
[4]it15047748@my.sliit.lk
[5]a.nugaliyadde@murdoch.edu.au
[6]yashas.m@sliit.lk

*Abstract*— Question Answering has recently received high attention from artificial intelligence communities due to the advancements in learning technologies. Early question answering models used rule-based approaches and moved to the statistical approach to address the vastly available information. However, statistical approaches are shown to underperform in handling the dynamic nature and the variation of language. Therefore, learning models have shown the capability of handling the dynamic nature and variations in language. Many deep learning methods have been introduced to question answering. Most of the deep learning approaches have shown to achieve higher results compared to machine learning and statistical methods. The dynamic nature of language has profited from the nonlinear learning in deep learning. This has created prominent success and a spike in work on question answering. This paper discusses the successes and challenges in question answering question answering systems and techniques that are used in these challenges.

## I. Introduction

Question Answering (QA) has been a challenging task in natural language understanding [1]. The key components in QA require the capability of understanding the question and the context in which the question is generated. QA has been deemed challenging due to dynamic nature of natural languages [1]. This has resulted in the application of data-driven methods in question answering. The idea is to allow the data, instead of the methods, do most of the work in question answering. This is due to a large number of text repositories that is available [2].

Rule-based approach was one of the initially used most prominent methods for QA systems. These systems utilized rules devised from grammatical semantics to determine the correct answer for a given question. These rules are usually handcrafted and heuristic, relying on lexical and semantic hints on context [3]. These rules exploit predefined patterns that classify questions based on the answer type. These grammatical rules represent the context in the form of decision trees and this was used to find the path that leads to the correct answer [4].

A major drawback of rule-based question answering systems was that the heuristic rules needed to be manually crafted. To devise these rules an in-depth knowledge of the semantics of a language was a necessity [5]. With the rapid growth of text material available online the importance of statistical approaches for QA has also increased. These approaches lean on predicting answers based on data. As these methods are capable of addressing the heterogeneity of data and free from structured query languages they have been adapted to various stages of QA [6].

Statistical approaches require the formation of a hypothesis before proceeding to build the model. This hypothesis sets the tone for the creation of the model. With the advancements in machine learning systems gained the capability to navigate the direction that the data dictates [7]. These inducted the self-learning capability to the QA systems. These systems are capable of building a knowledge base (taxonomy) from the training data it is provided and then use it to answer the actual questions. This brought a level of independence to the systems that were not quite there in rule-based or statistical approaches. Furthermore, as these systems are capable of optimizing itself over time it became one of the most lucrative approaches for QA [8].

Induction of Neural Networks for QA systems opened up a plethora of possibilities. Conventional machine-learning techniques were limited in their ability to process natural data in their raw form [38]. For decades, constructing a machine-learning system required careful engineering and considerable domain expertise to design a feature extractor that transformed the raw data into a suitable internal representation from which the learning subsystem, often a classifier, could detect or classify patterns in the input [38]. Deep learning methods are representation learning methods that allow a machine to be fed with raw data and to automatically discover the representations needed for detection, prediction or classification [38]. These are with multiple levels of representation, obtained by composing simple but non-linear modules that each transform the representation at one level (starting with the raw input) into a representation at a higher, slightly more abstract level [38]. With the composition of enough such transformations, very complex functions can be learned. For classification tasks, higher layers of representation amplify aspects of the input that are important for discrimination and suppress irrelevant variations [38].

Networks such as Dynamic Memory Networks [9], Reinforced-Memory Networks currently provide the state of the art results in Question Answering bringing Artificial Intelligence closer to human perception.

## II. RULE-BASED APPROACH

The initial question answering systems were logical representations of decision trees. The decision trees were linguistic structures that mirrored the way humans understand text based on grammatical rules. At the very beginning, all these rules were written by hand. These systems relied on the constant extension of functionality by the addition of rules that opened up new paths in the decision tree. QA systems based on such decision trees used lexical and semantic heuristics to see find evidence whether a sentence contains an answer to a question. These systems require rule sets that define paths for questions based on the question type. The path taken by the answer extraction process for "where is Tajmahal" would be different from "Who is on the one-dollar bill". To improve the answer matching syntactic analysis, morphological analysis, Part-Of-Speech tagging and Named Entity Recognition were later incorporated to these systems.

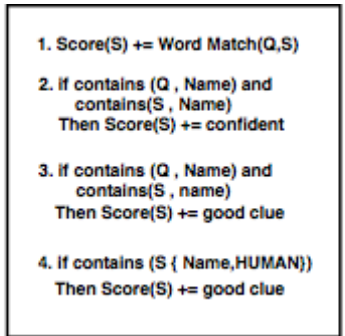

Fig. 1 Example of a rule-based algorithm

Though these systems were successful initiatives to the Question Answering domain they were not without flaws. Extensions to these systems required new rules to be introduced to the system which is cumbersome and makes these systems not suited for domains with highly volatile data [4].

As these new changes need to be explicitly programmed this slows down the development of the system considerably. These systems are highly reliant on the linguists who are creating the rules as these are unable to learn. These issues in the rule-based systems brought up the necessity for a self-learning approach that would solve the extensibility conundrum.

## III. STATISTICAL APPROACH

In the present research state, quick evolution in available online text repositories and web data has amplified the prominence of statistical approaches. These approaches put forward such techniques, which cannot only deal with the very large amount of data but their diverseness as well [6]. Additionally, statistical approaches are also independent of structured query languages and can formulate queries in natural language form. These approaches basically require a satisfactory amount of data for precise statistical learning but once properly learned, produce better results than other state-of-art approaches. Furthermore, the learned statistical program or method can be easily customized to a new domain being independent of any language form.

In general, statistical techniques have been so far successfully applied to the different stages of a QA system. Support vector machine (SVM) classifiers, Bayesian classifiers, Maximum entropy models are some techniques that have been used for question classification purpose. These statistical measures analyze questions for making a prediction about users' expected answer type. These models are trained on a corpus of questions or documents that have been annotated with the particular mentioned categories in the system [6].

One of the pioneer works based on the statistical model was IBM's statistical QA [10] system. This system utilized maximum entropy model for question/ answer classification based on various N-gram or bag of words features. And in IBM's statistical QA TREC-11 [11] they incorporated a novel extension of statistical machine translation. In this paper they have represented their model as p(c|a, q) which attempts to measure the c, 'correctness', of the answer and question And the paper introduced another hidden variable for the model e which represents the class of the answer. the new model is represented as

$$p(c|q, a) = P\ e\ p(c, e|q, a)$$
$$= P\ e\ p(c|e, q, a)p(e|q, a)$$

Further, in a different approach, Answer filtering via Text Categorization in Question Answering Systems [12] has been used Support Vector Machine text classifier for answer filtering. In this approach, document d is described as a vector

$d = <w_f^d, \ldots, w_{f_{|F|}}^d>$ in a |F|- dimensional vector space, where F is the adopted set of features.
$f_1, \ldots, f_{|F|} \in F$ set features from training document.
$w_f^d j \in F$ are document weights

$$w_f^d = \frac{l_f^d \times IDF(f)}{\sqrt{\sum_{r \in f}(l_f^d \times IDF(r))^2}}$$

$l_f^d$, the logarithm of the term frequency defined as:

$$l_f^d = \begin{cases} 0 & \text{if } o_f^d = 0 \\ \log(o_f^d) + 1 & \text{otherwise} \end{cases}$$

where the IDF(f) (the Inverse Document Frequency)

Stochastic approaches have brought about a significant improvement in POS tagging for the Sinhala Language [32]. This approach suggested a stochastic model that will be used in a statistical POS tagger to decide on the best suited tag for a particular word.

This system utilizes the statistical model for handling disambiguation, decoding and smoothing of unknown words. In Order to handle these three properties they used following techniques disambiguation - Hidden Markov Model (HMM), decoding - Viterbi Algorithm and for Smoothing - Linear Interpolation [32]. The tag which provides maximum of transition probabilities in combination with the previous two tags is selected as the tag for previously unseen tags.

Later, hybrid models which utilizes both rule-based and stochastic tagging approaches have been presented [33]. In this approach, a Hidden Markov Model based stochastic tagger is constructed initially. Later, since Sinhala is a morphologically rich language, rules defined on morphological features have been utilized to present the accurate and relevant tags for the given words.

## IV. MACHINE LEARNING APPROACH

With the introduction of machine learning to the QA domain algorithms that can learn to understand linguistic features without explicitly being told to. Statistical methods paved the path for this approach that enables the system to analyse an annotated corpus (training set) and then build a knowledge base. Context is processed usually by Named Entity Recognition techniques that act as the classifier to build a taxonomy [40]. This taxonomy then acts as the knowledge base. The questions would also be subjected to the same process. The major advantage machine learning brings to the table is its learnability. This makes the system highly scalable as long as there is enough training data.

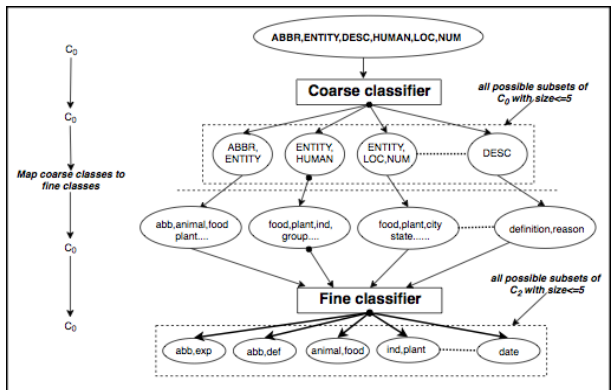

Fig. 2 Architecture of a machine learning context classifier

As most of the fields with a requirement for QA systems such as Journalism and Legal have a penchant for having large data machine learning approach has become quite popular in Question Answering. These systems regularly rake in state of the art results because of the speed and coverage factors that benefit because of the extensibility of this approach over the rule-based techniques.

Machine learning is combined with statistical approaches often in linguistic and sentiment related fields [34]. Successful systems have been developed using basic machine learning classifiers and strong underpinning feature sets [34]. with several word dictations. Systems developed with the strongest features and sentiment lexicons usually utilize linear support vector machines to train their underlying systems [34]. Similarly, an ensemble meta-classifier based approach, was able to successfully reproduce previous implementations, where it calculates the average confidence score of individual classifiers for positive, negative and neutral classes [34]. Ensemble approach has proven to be powerful in combining several methods.

## V. DEEP LEARNING APPROACH

Deep learning fundamentally differs from machine learning because of the ability it has to learn underlying features in data using neural networks. A standard neural network (NN) consists of many simple, connected units called neurons, each producing a sequence of real-valued activations [23]. Input neurons get activated through sensors perceiving the environment, other neurons get activated through weighted connections from previously active neurons [23]. Some neurons may influence the environment by triggering actions. Each input has a weight associated with it that depicts the importance of it compared to other inputs [23]. Recently, deep learning models have obtained a significant success on various natural language processing tasks, such as semantic analysis, machine translation and text summarization [24]. Neural Network architectures map textual context into logical representations that are then used for answer prediction. These neural networks utilize bi-directional Long Short-Term Memory(LSTM) units for question processing and answer classification [24].

Recurrent Neural Networks have shown promise in natural language processing. These networks differ from traditional neural networks because of the relationships the hidden layers maintain with the previous values [35]. This recurrent property gives RNN the potential to model long span dependencies [35] [39]. This allows the network to cluster similar histories that allows efficient representation of patterns with variable length [36]. However recurrent networks are not without its flaws. The major issue with Recurrent Neural Networks is that gradient computation becomes increasingly ill-behaved the farther back in time an error signal must be propagated, and that therefore learning arbitrarily long span phenomena will not be possible [36]. To address this issue several techniques have been introduced. Out of these Backpropagation through time(BPTT) is the most probable [37]. With BPTT, the error is propagated through recurrent connections back in time for a specific number of time steps [37]. Thus, the network learns to remember information for several time steps in the hidden layer when it is learned by the BPTT.

The Sinhala Question Answering System "Mahoshada" intends to obtain accurate information from a Sinhala tagged corpus [21]. It is a QA system for the Sinhala language which provides answers to any question within the context. "Mahoshada" has produced the capability to adopt any specified domain by the corpus provided. The system can take any annotated Sinhala text document under a specific domain [21]. The system summarizes the tagged corpus and utilizes it to generate answers for a given query. Summarized corpora are classified to enable fast retrieval of information. The system consists of four modules which are Document Summarizing, Document categorizing, Question processing, and Answer processing [21].

Document summarization is important to summarize multiple documents input by the user to reduce the number of terms and increase efficiency [21]. Document processing does the organization of the documents to retrieve answers conveniently by categorization. Question processing is responsible to analyse the question type and identify the question type. Answer processing involves in identifying and retrieving the most suitable answer [21].

Recently, Dynamic Memory Networks (DMNs) have shown success in Question Answering [9]. It is a neural network architecture that can process input sequences, form episodic memories and produce appropriate answers [13]. Questions posed by the user initiates an iterative attention process, where the model focuses attention on the inputs and previous iteration results [13]. DMN has achieved fine results on sentiment analysis and attained state-of-the-art results of the Facebook bAbI dataset [9][28]. Following is a brief overview of the modules of DMNs. DMN consists of four modules which are input module, question module, Episodic memory module and Answer module.
Input Module - Forms a distributed vector representation of the raw text input [13].

Question Module - Forms a distributed vector representation for the question of the task. The representation generated is then fed into the episodic memory module [13].
Episodic Memory Module – Episodic memory module decides which portion of inputs to focus on utilizing attention mechanisms given a collection of input representations [9] [13]. It then takes the question and previous memory into account and produces a "memory" vector representation.
Answer Module – Answer is generated from the final memory vector of the episodic memory module [13].

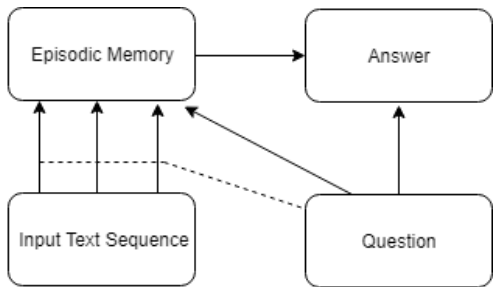

Fig. 4 Dynamic Memory Network

Another research that has been done is a framework, R-MN (Reinforced Memory Network) termed by combining Reinforcement Learning (Q-learning) and Memory Network (MN) to carry out QA tasks using Dynamic Memory Network (DMN) and Long Short-Term Memory (LSTM) [14]. Questions and input text sequences are taken as inputs to both MN and Q-Learning. The output of the MN is passed to Q-Learning as its second output for fine-tuning. This method has shown superior performance relative to DMN and other traditional neural networks with high accuracies [14]. The combination has improved the QA tasks utilizing less training data.

R-MN consists of three modules which are Input module, MN module and Q-Learning module [14].

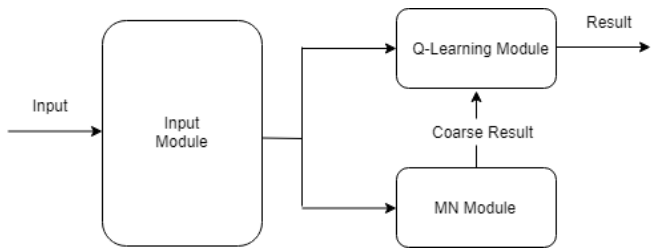

Fig. 5 Reinforcement Memory Network

## VI. NEURAL NETWORK MODELS

RNNs have been able to make a significant improvement in contrast to other machine learning techniques due to their ability to learn and carry out complicated transformations of data its ability maintaining long-term as well as short-term dependencies. They are said to be 'Turing-Complete' [16], therefore having the capacity to simulate arbitrary procedures. RNN contains an interplay of Reasoning, Attention, and Memory, commonly referred to as the 'RAM model' in Deep Neural Networks (DNN). In order to obtain state-of-the-art results different models have been introduced overtime in order to improve the performance of DNNs.

Researchers have been keen on building models of computation with various forms of explicit storage. Google's DeepMind project released Neural Turing Machines that extend the capabilities of neural networks by coupling them to external memory resources, which they can interact with by attentional processes [16]. NTM is similar to a regular computer with Von Neumann architecture that reads and writes to a memory. The significance is that these operations are differentiable and the controller that handles reads and writes is a neural network. It is shown that NTMs can infer simple algorithms such as copying, sorting, and associative recall from input and output examples. One of the main reasons why NTMs were so popular is that they are derived from psychology, cognitive science, neuroscience, and linguistics. Short-Term memory in NTMs is handled by resembling a working memory system, designed to solve tasks that require the application of approximate rules to data that are quickly bound to memory slots known as "rapidly-created variables". Additionally, NTMs use an attentional process to read from and write to memory selectively. Fig. 6 shows the architecture of a Neural Turing Machine.

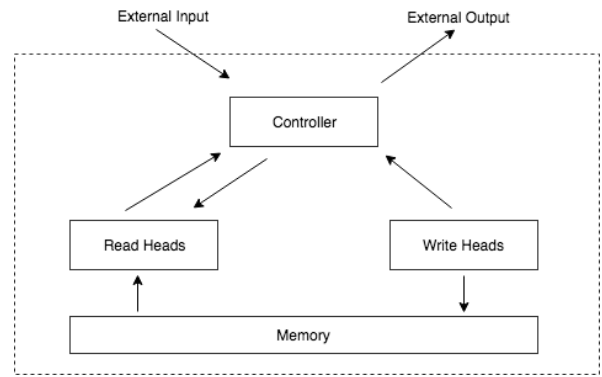

Fig. 6 Neural Turing Machine

End-to-end Memory Networks introduces a recurrent attention model over a possibly large external memory [20]. Since they are trained end-to-end it requires significantly less supervision during training. This technique is applicable to synthetic question answering and to language modelling. An RNN architecture is presented where it reads from a long-term memory multiple times before producing the output hence, improving its ability to reason. An RNN has a single chance to look at the inputs that are being fed one by one in order. But End-to-end memory network's architecture places all its inputs in the memory and the model decides which part to read next. An end-to-end memory network is demonstrated in Fig. 7.

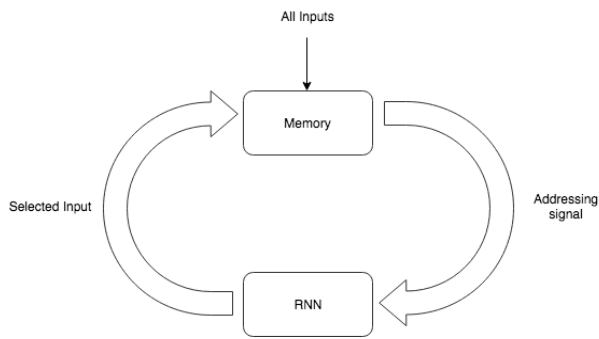

Fig. 7 End to End Memory Network

After NTMs, Facebook AI research introduced Stack-Augmented Recurrent Nets. They have attempted to show that some basic algorithms can be learned from sequential data using a recurrent network associated with a trainable memory [17]. Although machine learning has progressed over the years, with the scaling up of learning algorithms, alternative hardware such as GPUs or large clusters have been necessary. It is not practical with real-world applications. This approach has increased the learning capabilities of recurrent nets by allowing them to learn how to control an infinite structured memory.

Neural Training Machines were more expensive than previously considered due to the utilization of an external memory. Thus, Reinforcement Learning Neural Turing Machines (RLNTM) was introduced [18]. RLNTMs use a Reinforcement Learning Algorithm to train Neural Network that interacts with interfaces such as memory tapes, input tapes, and output tapes, to solve simple algorithmic tasks. RLNTMs use Reinforce algorithm to learn where to access the discrete interfaces and to use the backpropagation algorithm to determine what to write to the memory and to the output. RLNTMs have succeeded at problems such as copying an input several times to the output tape, reversing a sequence, and a few more tasks of comparable difficulty. Fig. 8 shows the architecture of a Reinforcement Learning Neural Turing Machine.

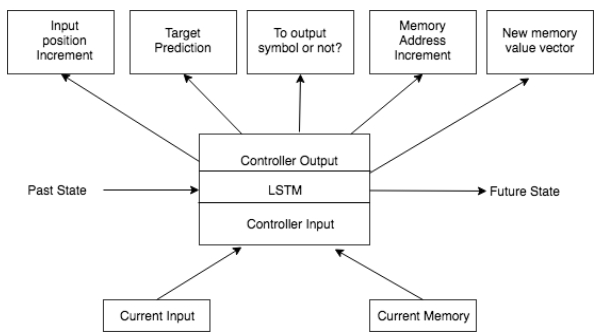

Fig. 8 Reinforcement Learning Neural Turing Machine

In addition to that, Neural Machine Translation is another approach to machine translation, which attempts to build and train a single large neural network that reads and outputs a correct translation instead of having small sub-components that are tuned separately [19]. Most machine translators are encoder-decoder models where the encoder neural network reads and encodes a sentence to a fixed length vector and the decoder output the translation from the encoded vector. The major issue with this system is that the neural network should be able to compress all the important information in the sentence to a fixed-length vector. Thus, dealing with longer sentences becomes difficult. The distinguishing feature of this approach is that it does not encode the entire input sentence into a fixed length vector. Instead, it encodes the input into a sequence of vectors and chooses a subset of these vectors adaptively while decoding [19]. This removes the potential challenge in dealing with long sentences and assists in retaining the necessary information.

A method has been proposed, utilizing a local attention-based model for Abstractive Sentence Summarization which up to date remains a challenge for Natural Language Processing. This model focuses on sentence-level summarization rather than using extracted portions of the sentences to prepare a condensed version. It uses a neural language model with an input encoder that learns a latent soft alignment over the input text to help inform the summary.

Although deep learning has achieved significant accomplishments, reasoning tasks remain elusive.

Neural networks have been applied in several languages specific question answering systems. Among these, a novel approach has been introduced to answer arithmetic problems using the Sinhala language [21]. A methodology is presented to solve Arithmetic problems in Sinhala Language using a Neural Network. This system comprises of keyword identification, question identification and mathematical operation identification and these are combined using a neural network [21]. Keywords are identified using Naïve Bayes Classification and questions are identified and mapped to the matching mathematical operator using Conditional Random operation which is performed on the identified keywords [21]. Sentence ordering is performed by the neural network according to "One vs. all Classification" [21]. All functions are combined through the neural network which builds an equation to solve the problem. This system learns to solve arithmetic problems with the accuracy of 76% [21].

VII. CONCLUSIONS

Though QA systems have evolved over the years they are still in need of improvements. QA systems have shown impressive accuracies on various domain based questions. However, commercial systems like IBM Watson, Google Assistant and Siri have shown impressive performance in QA but further improvements are required. Domain-independent QA, reasoning based QA, inferencing QA and complex QA requires many improvements. Many deep learning approaches have shown a potential of achieving these QA tasks but require many improvements. Handling domain independencies, reasoning, complex questions and inferencing from other information require more effective and learning models. This requires further analysis of the QA. However, there is a potential of moving out from deep learning and adapting the answering strategies from humans also show higher potential to achieve QA.

REFERENCES

[1] L. Kodra and E. Kajo, "Question Answering Systems: A Review on Present Developments, Challenges and Trends", *International Journal of Advanced Computer Science and Applications*, vol. 8, no. 9, 2017 [Online]. Available: https://thesai.org/Downloads/Volume8No9/Paper_31-Question_Answering_Systems_A_Review_on_Present_Developments.pdf. [Accessed: 22- May- 2018].

[2] E. Brill, J. Lin, M. Banko, S. Dumais and A. Ng, "Data-Intensive Question Answering", *Trec.nist.gov*, 2018. [Online]. Available: https://trec.nist.gov/pubs/trec10/papers/Trec2001Notebook.AskMSRFinal.pdf. [Accessed: 22- May- 2018].

[3] H. Madabushi and M. Lee, "High Accuracy Rule-based Question Classification using Question Syntax and Semantics", *Aclweb.org*, 2018. [Online]. Available: http://www.aclweb.org/anthology/C16-1116. [Accessed: 23- May- 2018].

[4] E. Riloff and M. Thelen, "A Rule-based Question Answering System for Reading Comprehension Tests", 2018. [Online]. Available: https://pdfs.semanticscholar.org/4454/06b0d88ae965fa587cf5c167374ff1bbc09a.pdf. [Accessed: 23- May- 2018].

[5] S. Humphrey and A. Brownea, "Comparing a Rule Based vs. Statistical System for Automatic Categorization of MEDLINE Documents According to Biomedical Specialty", 2018. [Online]. Available: https://www.ncbi.nlm.nih.gov/pmc/articles/PMC2782854/. [Accessed: 23- May- 2018].

[6] S. K. Dwivedia and V. Singh, "Research and reviews in question answering system," in *Proceedings of International Conference on Computational Intelligence: Modeling Techniques and Applications*, 2013, pp. 417 – 424 .

[7] D. Cohn, Z. Ghahramani and M. Jordan, "Active Learning with Statistical Models", *Journal of Artificial Intelligence Research*, vol.4, 1996. [Online]. Available: https://jair.org/index.php/jair/article/view/10158. [Accessed: 22- May- 2018].

[8] X. Li and D. Roth, "Learning Question Classifiers", *Dl.acm.org*, 2018. [Online]. Available: https://dl.acm.org/ft_gateway.cfm?id=1072378&ftid=569844&dwn=1&CFID=37296624&CFTOKEN=52eb69472489ceca-89B68F45-9568-FB11-CC44151C22772CA6. [Accessed: 22- May- 2018].

[9] Raghuvanshi, A., & Chase, P., "Dynamic Memory Networks for Question Answering". [Online]. Available: https://cs224d.stanford.edu/reports/RaghuvanshiChase.pdf. [Accessed: 22- May- 2018].

[10] Ittycheriah A., Franz M, Zhu WJ, Ratnaparkhi A. and Mammone R. J., "IBM's statistical question answering system," in *Proceedings of the Text Retrieval Conference TREC-9*, 2000.

[11] Ittycheriah, A., & Roukos, S. (2006). "IBMs Statistical Question Answering System" - TREC-11. doi:10.21236/ada456310

[12] Moschitti A. "Answer filtering via text categorization in question answering systems," in *Proceedings of the 15th IEEE International Conference on Tools with Artificial Intelligence*, 2003, pp. 241-248.

[13] Kumar, A., Irsoy, O., Iyyer, M., Ondruska, P., Bradbury, J., Gulrajani, I., Socher, and R. (n.d.). "Dynamic Memory Networks for Natural Language Processing", [Online]. Available: http://proceedings.mlr.press/v48/kumar16.pdf. [Accessed: 22- May- 2018].

[14] A. Nugaliyadde, Kok Wai, WongFerdous, and SohelHong Xie. "Reinforced Memory Network for Question Answering," in *Proceedings of International Conference on Neural Information Processing*, 2017. [Online]. Available: https://www.researchgate.net/publication/320658588_Reinforced_Memory_Network_for_Question_Answering. [Accessed: 22- May- 2018].

[15] Siegelmann, H. T. and Sontag, E. D, "On the computational power of neural nets", *Journal of computer and system sciences*, vol. 50, 1995. [Online]. Available: https://www.sciencedirect.com/science/article/pii/S0022000085710136. [Accessed: 22- May- 2018].

[16] Alex Graves, Greg Wayne and Ivo Danihelka. (2014) "Neural Turing Machines". [Online]. Available: https://arxiv.org/pdf/1410.5401v2.pdf. [Accessed: 22- May- 2018].

[17] Armand Joulin Facebook and Tomas Mikolov. (2015) "Inferring Algorithmic Patterns with Stack-Augmented Recurrent Nets". [Online]. Available: https://arxiv.org/pdf/1503.01007.pdf. [Accessed: 23- May- 2018].

[18] Wojciech Zaremba, and Ilya Sutskever. (2016) "REINFORCEMENT LEARNING NEURAL TURING MACHINES". [Online]. Available: https://arxiv.org/pdf/1505.00521.pdf. [Accessed: 23- May- 2018].

[19] Dzmitry Bahdanau, KyungHyun Cho and Yoshua Bengio. (2014) "NEURAL MACHINE TRANSLATION BY JOINTLY LEARNING TO ALIGN AND TRANSLATE". [Online]. Available: https://arxiv.org/pdf/1308.0850.pdf. [Accessed: 22- May- 2018].

[20] Sainbayar Sukhbaatar, Arthur Szlam, Jason Weston and Rob Fergus (2015). "End-To-End Memory Networks". [Online]. Available: https://arxiv.org/pdf/1503.08895.pdf. [Accessed: 22- May- 2018].

[21] T. Jayakody, T.S.K. Gamlath, W.A.N. Lasantha, K.M.K.P. Premachandra, Y. Mallawaarachchi, and A. Nugaliyadde, "Mahoshadha, The Sinhala Tagged Corpus based Question Answering System," in *Proceedings of International Conference on Information and Communication Technologies for Intelligence Systems, 2015*. [Online] Available: https://www.researchgate.net/publication/292615736_Mahoshadha_The_Sinhala_Tagged_Corpus_based_Question_Answering_System. [Accessed: 23- May- 2018].

[22] M. Chathurika, C. De Silva, A. Raddella, E. Ekanayake, Y. Mallawarachchi and A. Nugaliyadde, "Solving Sinhala Language Arithmetic Problems using Neural Networks," in *Proceedings of 34th National Information Technology Conference*, 2016. [Online]. Available: https://www.researchgate.net/publication/305262356_Solving_Sinhala_Language_Arithmetic_Problems_using_Neural_Networks. [Accessed: 22- May- 2018].

[23] J. Schmidhuber, "Deep learning in neural networks: An overview", 2018. [Online]. Available: https://ac.els-cdn.com/S0893608014002135/1-s2.0-S0893608014002135-main.pdf?_tid=7f19f037-5d16-4521-9163-a8b2a8f3768c&acdnat=1527478744_44e66fda06319991375c434ae2fc5b60. [Accessed: 23- May- 2018].

[24] M. Tan, C. Santos, B. Xiang and B. Zhou, "LSTM-based Deep Learning Models for Non-factoid Answer Selection", 2015. [Online]. Available: https://arxiv.org/abs/1511.04108. [Accessed: 24- May- 2018].

[25] B. Magnini, D. Giampiccolo, P. Forner, C. Ayache, V. Jijkoun, P. Osenova, A. Peñas, P. Rocha, B. Sacaleanu and R. F. E. Sutcliffe, "Overview of the CLEF 2006 Multilingual Question Answering Track," in *Proceedings of Evaluation of Multilingual and Multi-modal Information Retrieval, 7th Workshop of the Cross-Language Evaluation Forum, CLEF 2006, Alicante, Spain*, 2006, pp. 223–256.

[26] A. Rodrigo and A. Peñas, "A Study about the Future Evaluation of Question-Answering Systems", 2017. [Online]. Available: https://www.researchgate.net/publication/319617514_A_Study_about_the_Future_Evaluation_of_Question-Answering_Systems. [Accessed: 24- May- 2018].

[27] A. Peñas, P. Forner, R. Sutcliffe, A. Rodrigo, C. Forăscu, I. n. Alegria, D. Giampiccolo, N. Moreau and P. Osenova, "Overview of ResPubliQA 2009: Question Answering Evaluation over European Legislation," in *Proceedings of the 10th cross-language evaluation forum conference on Multilingual information access evaluation: text retrieval experiments*, pp. 174–196, 2009.

[28] J. Weston, A. Bordes, S. Chopra, T. Mikolov, "Towards AI-Complete Question Answering: A Set of Prerequisite Toy Tasks", 2016. [Online]. Available: https://arxiv.org/abs/1502.05698. [Accessed: 24- May- 2018].

[29] D. A. Ferrucci, E. W. Brown, J. Chu-Carroll, J. Fan, D. Gondek, A. Kalyanpur, A. Lally, J. W. Murdock, E. Nyberg, J. M. Prager, N. Schlaefer, and C. A. Welty, "Building Watson: An Overview of the DeepQA Project", AI Magazine 31 (3), 2010, pp. 59–79.

[30] S. Harabagiu, F. Lacatusu, and A. Hickl, "Answering Complex Questions with Random Walk Models," in *Proceedings of the 29th Annual International ACM SIGIR Conference on Research and Development in Information Retrieval*, SIGIR '06, 2006, pp. 220–227.

[31] E. Saquete, P. Martínez-Barco, R. Muñoz, and J. L. Vicedo, "Splitting Complex Temporal Questions for Question Answering Systems," in *Proceedings of the 42nd Annual Meeting on Association for Computational Linguistics, ACL '04*, 2004.

[32] Jayasuriya, M., and Weerasinghe, A. R. "Learning a stochastic part of speech tagger for sinhala," in *Proceedings of the 2013 International Conference on Advances in ICT for Emerging Regions (ICTer)*, 2013. doi:10.1109/icter.2013.6761168

[33] Gunasekara, D., Welgama, W., and Weerasinghe, A. "Hybrid Part of Speech tagger for Sinhala Language," in *Proceedings of the 2016 Sixteenth International Conference on Advances in ICT for Emerging Regions (ICTer)*, 2016. doi:10.1109/icter.2016.7829897

[34] Muthutantrige, S. R., and Weerasinghe, A. "Sentiment Analysis in Twitter messages using constrained and unconstrained data categories," in *Proceedings of the Sixteenth International Conference on Advances in ICT for Emerging Regions (ICTer)*, 2016. doi:10.1109/icter.2016.7829935

[35] T. Mikolov and G. Zweig, "Context Dependent Recurrent Neural Network Language Model", 2012. [Online]. Available: https://www.microsoft.com/en-us/research/wp-content/uploads/2012/07/rnn_ctxt_TR.sav_.pdf. [Accessed: 24- May- 2018].

[36] T. Mikolov, S. Kombrink, L. Burget, J. Cernocký and S. Khudanpur, "Extensions of recurrent neural network language

model," in *Proceedings of IEEE International Conference on Acoustics, Speech and Signal Processing (ICASSP)*, 2011.

[37] R. Williams and D. Zipser, "A Learning Algorithm for Continually Running Fully Recurrent Neural Networks", *Neural Computation*, 1, pp. 270-280, 1989.

[38] Y. Bengio and G. Hinton, "Deep learning", 2015. [Online]. Available: https://www.nature.com/articles/nature14539.

[39] Boukoros, Spyros, Anupiya Nugaliyadde, Angelos Marnerides, Costas Vassilakis, Polychronis Koutsakis, and Kok Wai Wong. "Modeling server workloads for campus email traffic using recurrent neural networks." In International Conference on Neural Information Processing, pp. 57-66. Springer, Cham, 2017.

[40] Senevirathne, K. U., N. S. Attanayake, A. W. M. H. Dhananjanie, W. A. S. U. Weragoda, A. Nugaliyadde, and S. Thelijjagoda. "Conditional Random Fields based named entity recognition for sinhala." In 2015 IEEE 10th International Conference on Industrial and Information Systems (ICIIS), pp. 302-307. IEEE, 2015.